\documentclass{article}

\usepackage{PRIMEarxiv}

\usepackage[utf8]{inputenc} % allow utf-8 input
\usepackage[T1]{fontenc}    % use 8-bit T1 fonts
\usepackage{hyperref}       % hyperlinks
\usepackage{url}            % simple URL typesetting
\usepackage{booktabs}       % professional-quality tables
\usepackage{amsfonts}       % blackboard math symbols
\usepackage{nicefrac}       % compact symbols for 1/2, etc.
\usepackage{microtype}      % microtypography
\usepackage{lipsum}
\usepackage{fancyhdr}       % header
\usepackage{graphicx}       % graphics
\graphicspath{{media/}}     % organize your images and other figures under media/ folder

\title{A Machine Learning Method for Predicting Traffic Signal Timing from Probe Vehicle Data
}

\author{
  Juliette Ugirumurera \\
  National Renewable Energy Laboratory\\
   \And
  Joseph Severino\\
  National Renewable Energy Laboratory \\
  \And
  Erik A. Bensen\\
  Carnegie Mellon University \\
  \And
  Qichao Wang\\
  National Renewable Energy Laboratory \\
  \And
  Jane Macfarlane\\
  UC Berkeley \\
}

\begin{document}
\maketitle

\begin{abstract}
Traffic signals play an important role in transportation by enabling traffic flow management, and ensuring safety at intersections. In addition, knowing the traffic signal phase and timing data can allow optimal vehicle routing for time and energy efficiency, eco-driving, and the accurate simulation of signalized road networks. In this paper, we present a machine learning (ML) method for estimating traffic signal timing information from vehicle probe data. To the authors best knowledge, very few works have presented ML techniques for determining traffic signal timing parameters from vehicle probe data. In this work, we develop an Extreme Gradient Boosting (XGBoost) model to estimate signal cycle lengths and a neural network model to determine the corresponding red times per phase from probe data. The green times are then be derived from the cycle length and red times. Our results show an error of less than 0.56 sec for cycle length, and red times predictions within 7.2 sec error on average.
\end{abstract}

% keywords can be removed

\section{INTRODUCTION}

In traffic networks, traffic signals play a key role in determining and managing vehicular traffic flow. They control the flow of traffic, ensure safety by regulating the flow of competing movements through intersections and can reduce traffic congestion when optimized \cite{guler2014using,rafter2020augmenting}. However, they can also induce stop-and-go traffic, which increases vehicles' delay and fuel consumption. In addition, not knowing the traffic signals' timings and switching patterns makes it very challenging to accurately determine the most time-efficient or energy-efficient routes, to inform driving decisions to maximize arrival on green or minimize engine idling at intersections \cite{fayazi2014traffic}, and to correctly simulate traffic for signal-controlled regions.

Usually, traffic lights are managed by different local agencies. For example, in the US, hundreds of agencies are in charge of the more than 320,000 traffic lights installed \cite{WinNT}. This makes direct access to traffic signal timing and phase data a very challenging task. In this paper, we present a machine learning method that uses probe vehicle data to estimate the timings for pre-timed traffic signals. We demonstrate this algorithm on probe data generated from a well-calibrated microscopic simulation using the SUMO simulator\cite{SUMO2018} and the NEMA Type controller in SUMO \cite{Wang2021}. We focus on fixed-time traffic signals since they represent the majority of traffic lights in the US \cite{chajka2021intelligent}.

\section{RELATED WORKS}
Probe vehicle data has enabled many important tasks in transportation research, including estimating link-level hourly traffic volume \cite{hou2018network}, queue length estimation \cite{comert2011analytical}, and traffic signal control \cite{lian2021adaptive}. 
The availability of and opportunities to exploit probe vehicle data will only continue to grow due to the predicted increase in connected and automated vehicle (CAV) adoption and the increased availability of high-speed communication infrastructure.
Probe vehicle data has also been used broadly for estimating traffic signal timings. In \cite{fayazi2014traffic}, the authors uses statistical patterns in probe data from public buses in San Fransisco, California, USA, to estimate the cycle times, the duration of red times and the greens start times for fixed-timed traffic lights. Yu and Lu estimated cycle length for fixed time intersections using low frequency taxi trajectory data\cite{yu2016learning}. They formulate the cycle length estimation problem as a general approximate greatest common divisor (AGCD) problem and solved it with a most frequent AGCD algorithm. The work in \cite{kerper2012learning} presents an optimization-based traffic signal state estimation algorithm on vehicle speed profiles to determine traffic signal phase schedules. Wang and Jiang \cite{wang2012traffic} determine the cycle length and green times of signalized intersection by analyzing the velocity-time curves of all vehicles going through the intersection. Chuan et al. calculated cycle length and green times using shockwave theory \cite{chuang2014discovering}. Finally, \cite{mahler2012reducing} and \cite{mahler2014optimal} focus on using probabilistic prediction of traffic signal timings. 

To our best knowledge, there are only a few works that study machine learning (ML) methods to estimate signal timing parameters from probe vehicles. Unlike mathematical methods, such as statistical, optimization and probabilistic based-methods, which require the explicit mathematical representation of the desired problem to solve, ML techniques can derive models from data to produce meaningful insights and predictions\cite{Edward.2021}. 
Advances in sensing and communication have lead to unprecedentedly large amounts of data, which, when combined with ML's spontaneous model-learning capability, explains the recent surge in ML applications across domains from health care to self-driving cars and traffic control.
For example, the work in \cite{protschky2015learning} uses a Bayesian learning approach to determine the cycle length from historic trajectory data, but without estimating the red or green times per phase. In \cite{genser2020time}, a Random Survival Forest and a Long-Short-Term-Memory (LSTM) neural network are used to predict the residual time per intersection phase; however, they do not estimate the cycle length or the green and red phase times; additionally, these models require vehicle detection data as input. In \cite{eteifa2021predicting}, the authors presents a deep learning based LSTM methodology to predict actuated traffic lights' switching time from green to red and vice versa, but require signal timing parameters as input rather than estimating them. 

In this work, we showcase the potential of ML for estimating pre-timed traffic signal timing by using an Extreme Gradient Boosting (XGBoost) model \cite{chen2016xgboost} to determine the cycle length and a dense neural network model \cite{anderson1995introduction} to estimate the red times from probe vehicle data. We choose XGBoost for estimating cycle length because it has been known to perform well on tabulated data sets \cite{shwartz2022tabular}. For predicting the red times, we tried several simpler models like XGBoost, Random Forest and Logistic Regression models, but none performed as well as the DNN due to the feature complexity. Traffic signals' green times are obtained by subtracting the red times from the cycle length for each intersection's phase. The yellow times are counted as part of the green times. Hence, our paper's contributions can be summarized as follows:
\begin{itemize}
  \item Development of a novel ML-based method for estimating pre-timed traffic signal timing parameters including cycle length, red times and green times, that is applicable to any location that has high-fidelity vehicle probe data.
  \item Demonstration of the developed ML-based method on a sizeable realistic network with 39 intersections, with results showing an error of less than 0.56 sec for cycle length, and red times predictions within 7.2 sec error on average. 
\end{itemize}

\section{METHODOLOGY}
\subsection{Simulation Description}
\begin{figure}[h!]
\centering
\includegraphics[width=6in]{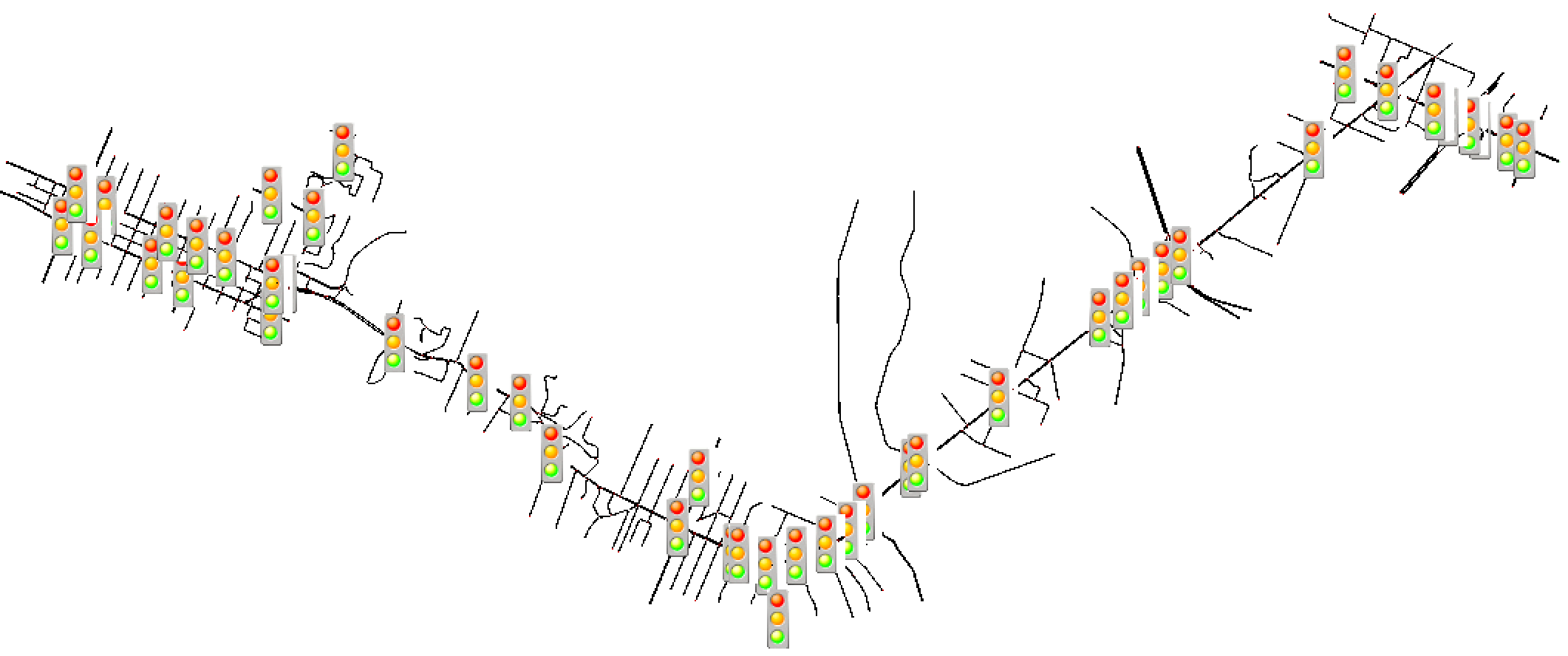}
\caption{Road network in Chattanooga, Tennessee, with 39 intersections used for traffic signal timing estimation}
\label{fig:chat_net}
\end{figure}

\begin{figure}[h!]
\centering
\includegraphics[width=6in]{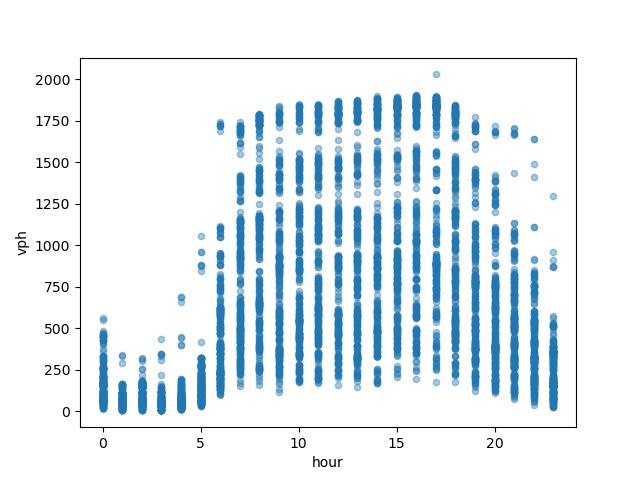}
\caption{Hourly historic, link-level, traffic volume estimates in vehicle per hour (vph) for the area of interest in Chattanooga, Tennessee, used for demand calibration}
\label{fig:chat_demand}
\end{figure}

In order to train and evaluate our ML models, we use simulated probe data from a realistic road network consisting of 39 intersections in the city of Chattanooga in Tennessee, USA, shown in Figure \ref{fig:chat_net}. The network geometry was validated using satellite images of the real-world network and its demand was calibrated using historic, link-level, traffic volume estimates \cite{hou2018network}, as shown in Figure \ref{fig:chat_demand}. The City of Chattanooga's department of transportation also provided signal timing information for the 39 intersections, which served as the ground truth for our models.

We used SUMO \cite{SUMO2018} to run our simulation for a day. SUMO gives us the advantage of controlling and observing all aspects of the traffic and signal timings. This eliminates any uncertainty that might be observed in the field when using signal timing sheets that might be out of date or data that is difficult to process because various formats. Additionally, SUMO provides a second by second record of vehicle movements via the FCD (floating car device) listener. Thus, we could store all vehicle trajectories in simulation and their speeds for each simulation run. We also ran simulations with different percentiles of the traffic demand in Figure \ref{fig:chat_demand} (for example, we ran simulation with 50\% of the hourly demand) to simulate demand variability. After running the simulations with various demand levels, we recorded FCD data for up to 50\% of the vehicles. We then aggregated each trip and filtered only trips that served our purpose of estimating cycle lengths, red times and green times. 

\subsection{Data Processing}
To process the data for feature engineering, we used the FCD trajectory data as inputs and the Traffic Light Scheme settings (tls) data as our target variable. The first step of processing the FCD data was to create a bounding box around each signalized intersection and filter out all the trajectories that passed within the bounding box. We set the bounding box to be 500 feet from the center of the intersection. This gave us enough room to develop aggregated metrics of each trajectory journeying through the intersection. After extracting the data points around each intersection, we then filtered only trips that stopped for any given length of time before the intersection and traversed through the center of the intersection with any vehicle movement and approach. 

\begin{figure}[h!]
\centering
\includegraphics[width=5in]{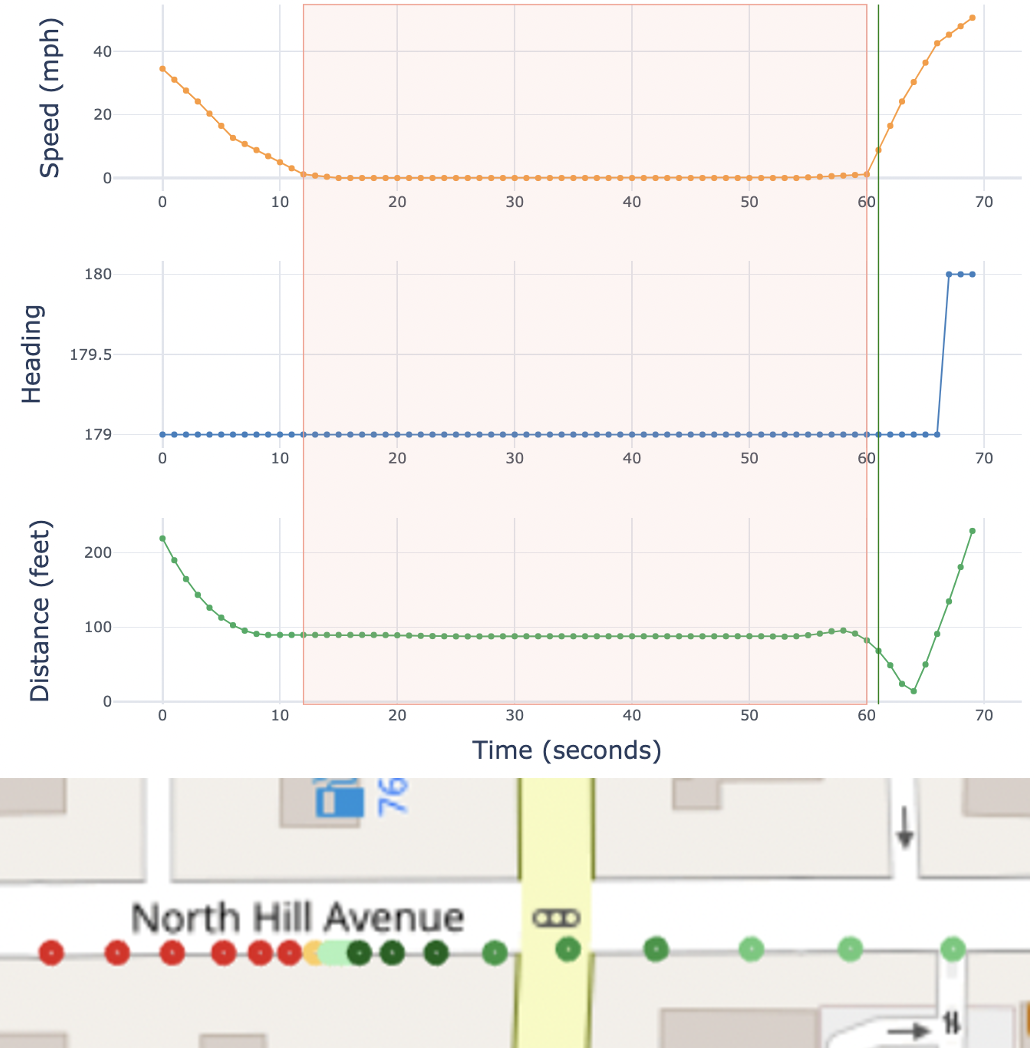}
\caption{Vehicle trajectory near an intersection, including a speed profile, a heading profile, distance profile, and at the bottom, the actual way points of the trajectory}
\label{fig:trip_metric}
\end{figure}

In the top panel of Figure \ref{fig:trip_metric}, we see one trip that approaches an intersection and stops for a duration of approximately 48 seconds shown by the shaded area in the top graphs. We can also see the way-points in the bottom panel showing the deceleration in red and acceleration in green of the vehicle before, through, and after the center of the signalized intersection. We filtered trips using four criteria. These criteria were originally built using real-world probe trajectories. Thus, this method translates easily to real-world probe data for future implementation. 

The first criteria is to find a pattern of deceleration, zero speed and acceleration of a vehicle trip. We select this pattern because if the vehicle never stops, we are unable to accurately estimate the initial onset of a green light given by a vehicle accelerating from zero. Thus, if a vehicle just passes through a light, we can only say the signal was green or yellow for the few seconds that the vehicle approaches and passes through the center of the intersection. This would not be sufficient information if we assume a low penetration rate of probe vehicles. 

The second criteria is based on sparsity of trace way points. For the simulation, this criteria did not matter since all data was recorded without sensing error. However, real-world data contain trips with missing way-points. Thus, we dropped any vehicle trip that was missing data for a period of 10 seconds or more. 

Third, we wanted to exclude any trips that were within 500 feet of the intersection, but did not pass through the center of the intersection. In the real-world, vehicles may be visiting a location at the corner of the intersection, such as a gas station or fast food restaurant, but never actually traverse the intersection. Thus, we only include vehicles that have a 50 feet minimum distance away from the center of the intersection. You can see in the top panel of Figure \ref{fig:trip_metric} on the Distance graph that the minimum distance can be easily calculated for each trip. 

The last criteria we filter on is total duration of trip. Considering that the majority of signalized intersections have a cycle length under 120 seconds, we exclude any trip with a duration of over 2 minutes around the intersection. With real-world probes, we came across some trips where the vehicle turned around after passing the intersection and passed back through the light. These types of trips were rare but added complexity and noise to our processing, and this required extracting out these acceleration times. 

We use acceleration times to estimate cycle lengths since, given enough vehicle data, there will be a periodicity between consecutive green lights if the signal timings are fixed. 
To get the acceleration start times, we use a simple algorithm that looks at each vehicle trip and first finds the first positive change in speed after the vehicle has stopped. This timestamp is then recorded and assigned to an intersection and a phase so that it can be later matched to the ground truth cycle length for a given hour. 

\subsection{Cycle Length Feature Engineering}
To create ML features from the acceleration start times, we hypothesized that the largest magnitude Fourier Frequencies from the Fourier Transform of the distribution of acceleration start times will be a good predictor of cycle lengths. This is because the change in signal phases throughout a cycle should cause vehicles to start accelerating at periodic intervals. To create these Fourier input features for the XGBoost model, we first group the acceleration start times by the intersection they were located at, the signal phase of the vehicle trajectory, and the time of day. This grouping ensures that all acceleration start times come from vehicles passing through a specific intersection with the same target cycle length. After binning, we perform the following to create the input features:
\begin{enumerate}
    \item Bin all start times within the first grouping into 1 hour time windows.
    \item For each one hour window perform the following:
    \begin{enumerate}
        \item Convert the date-time representation of each start time to the seconds after the first start time within the current time window.
        \item Approximate the distribution of start times within the current time window using a Gaussian Kernel Density Estimator (KDE) with bandwidth 6.
        \item Take the fast Fourier Transform (fft) of the KDE and save the 30 frequencies that had the largest magnitude of their Fourier amplitudes.
    \end{enumerate}
    \item Repeat for all groupings
\end{enumerate}

In doing this process, one data point for training and testing the ML model consists of information derived from a 1 hour time window of vehicles moving through a specific intersection. We chose the bandwidth (standard deviation) of the Gaussian KDE to be 6 seconds so that it was large enough to ensure starts within the same cycle blend together. Second, we saved the top 30 most significant Fourier frequencies arbitrarily; however, we optimize the number of frequencies used by the model along with the other XGBoost hyperparameters. Finally, we discard all time windows with less than two starts because having no starts in a window provides no information and only one start in a window only provides artificial noise in the Fourier frequencies from the length of the time window and the frequencies that compose the single Gaussian in the KDE. We also optimize and analyze the minimum number of starts per hour as part of a multistage tuning process. Finally, to normalize the inputs, we use the StandardScalar transformation provided by Sci-kit Learn\cite{sklearn}.

\subsection{Cycle Length Training Methods}

\begin{table}[tb]
    \centering
    \caption{Optimized hyperparameters for the XGBoost model.}
%Optimization used the following fixed parameters: booster = gbtree and grow\_policy = depthwise.    
    \label{table:opt_values}
    \begin{tabular}{r||l|l}
         Parameter & Search Space & Optimal Value \\ \hline \hline
         n\_estimators & $100$ \textendash\ $2000$ & $2000$\\ 
         learning\_rate & $10^{-4}$ \textendash\ $10^{0}$ & $10^{0.0}$ \\ 
         max\_depth & $2$ \textendash\ $20$ & $17$ \\ 
         gamma & $10^{-5}$ \textendash\ $10^{0}$ & $10^{-5.0}$ \\ 
         min\_child\_weight & $1$ \textendash\ $10$ & $1.499$ \\ 
         subsample & $0.5$ \textendash\ $1.0$ & $1.0$ \\ 
         colsample\_by\_tree & $0.5$ \textendash\ $1.0$ & $1.0$ \\ 
         n\_fourier & $2$ \textendash\ $30$ & $1$ \\ 
         min starts & $2$ \textendash\ $250$ & 20\\ \hline 
    \end{tabular}
\end{table}

For training the XGBoost model, we first randomly selected $20\%$ of the points to set aside as the test set. Then, we randomly divide the training dataset into 5 folds for cross validation (CV), which is used during the optimization process. To optimize the XGBoost hyperparameters, the number of Fourier frequencies and minimum starts per hour, we used the following multistage optimization process:
\begin{enumerate}
    \item Optimize the hyperparameters and number of Fourier frequencies using Bayesian Optimization from the bayes\_opt package in Python\cite{bayesopt} with 10 initialization points and 50 iterations.
    \item Compute the CV predictions using the optimized hyperparameters and use them to compute the MAE of all points with more than $c$ acceleration starts for $c=2,3,...,249,  250$.
    \item Choose the optimal cutoff to be the minimum value of $c$ such that the associated fraction of CV predictions with an error less than 2 seconds is at least $95\%$.
    \item Down select the training and testing data to only include points with $>c$ acceleration starts and re-optimize the hyperparameters using the same Bayesian Optimization scheme.
\end{enumerate}

\subsection{Red Times Feature Engineering}
To predict the red times, we start with the vehicle stop durations instead of the acceleration start times used for cycle estimation. We bin the vehicle stop times by intersection, direction of travel (North, South, East, West) and time of day (AM, PM, Off Peak). With this binning strategy, all vehicle stop times in a bin correspond to the same unique phase and target red times, and this means we can incorporate distributional information about the stop times into the input features.

Next, we perform up-sampling to improve the granularity of the examples and evenly
4 weight all target values in the training process. To do this, we create groupings of vehicle stop times from each bin by sampling 50 stop times with replacement from each bin to create a grouping and repeating this 40 times. The sampling ensures empirical cumulative distribution functions (described below) were created from 50 stop times on average and the 40 repetitions presents distinct permutations of the data to the network for training and validation. This is useful to improve network interpolation between training points. We chose 40 repetitions to balance the increased number of unique permutations with the training time of the network.

We can then encode the distributional information as input features by using empirical quantiles. Let $s_i$ be the stop time of the $i^{th}$ vehicle, $\alpha$ be the quantile of interest, and $n$ be the number of vehicle stop times in the grouping. Then the $\alpha$ quantile $Q_\alpha$ is given by Equation \ref{eq:quantile}. 
\begin{equation}\label{eq:quantile}
    Q_\alpha = \max \left\{s_i\ \Bigg| \  \frac{\# \left\{ s_j | s_j \leq s_i \right\}}{n} < \alpha \right\}
\end{equation}
Where $\{ \}$ denotes a set, the notation $\{ s_i\ |\ X\}$ means the set of $s_i$ values that satisfy condition $X$ and $\# A$ indicates the number of elements in the set $A$. We use a $100\times1$ vector of the $1\%, 2\%, ..., 99\%, 100\%$ quantiles as the input to the neural network. Finally, we also use the StandardScalar transform from Sci-kit Learn\cite{sklearn} to normalize the inputs before training. Note that the $100\%$ quantile is just the maximum vehicle stop time from that grouping.

\begin{figure}[h!]
\centering
\includegraphics[width=6in]{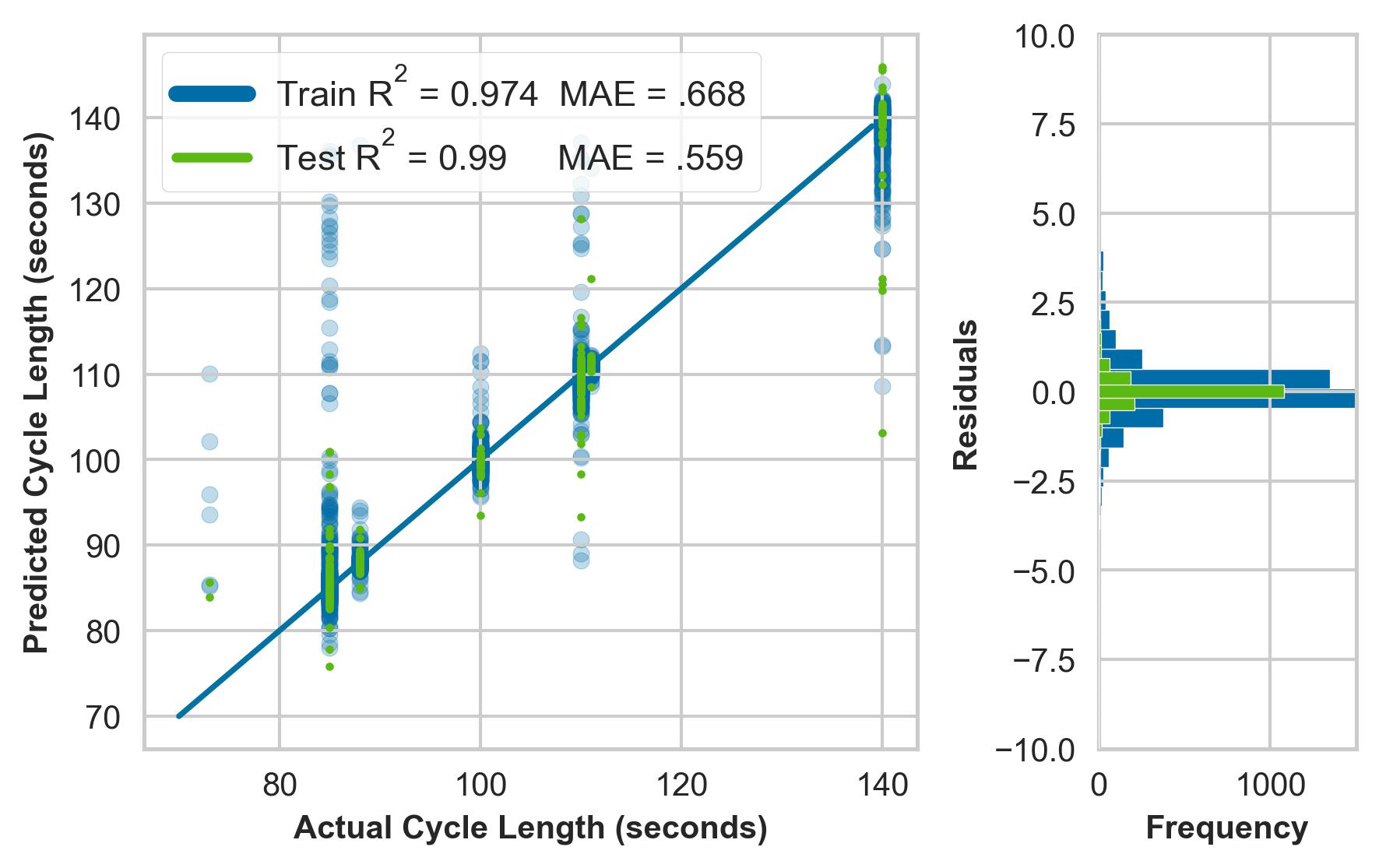}
\caption{Cycle Length Parity Plot}
\label{fig:cycle_lengths}
\end{figure}

\subsection{Red Times Training Methods}
To predict the red times, we use a Dense Neural Network (DNN). The DNN inputs a $100\times 1$ vector, outputs a scalar prediction and has 11 hidden layers of sizes: 550, 1000, 900, 800, 700, 600, 500, 400, 300, 200, 100 respectively. All layers used a Leaky ReLU activation function with $\alpha=0.01$. To train the model, we divided the data randomly into a 20\% validation set and 80\% training set. Then we trained the model with minibatch size set to 32 until we reached an early stopping trigger with patience set to 50. The Neural Network was constructed using the Tensorflow\cite{tensorflow} and Keras\cite{keras} packages in Python.

\section{RESULTS AND DISCUSSION}
\subsection{Cycle Length Results}
After the first round of the multi step optimization routine for the XGBoost model, we determined the minimum starts points per hour cutoff to be 20. We chose this cutoff because it is the minimum number of starts per hour for the model to produce 95\% of it's estimations with an error less than 2 seconds. Table \ref{table:opt_values} shows the results from the final Bayesian Optimization to optimize the hyper parameters and number of Fourier frequencies.

The results for the cycle length model have a mean absolute error (MAE) less than 0.7 seconds for both training and testing sets, along with a R-squared values of .97 and .99 for training and testing respectively. These results are shown in figure \ref{fig:cycle_lengths} with green being the test set and blue being the training set. 

\begin{figure}[h!]
\centering
\includegraphics[width=6in]{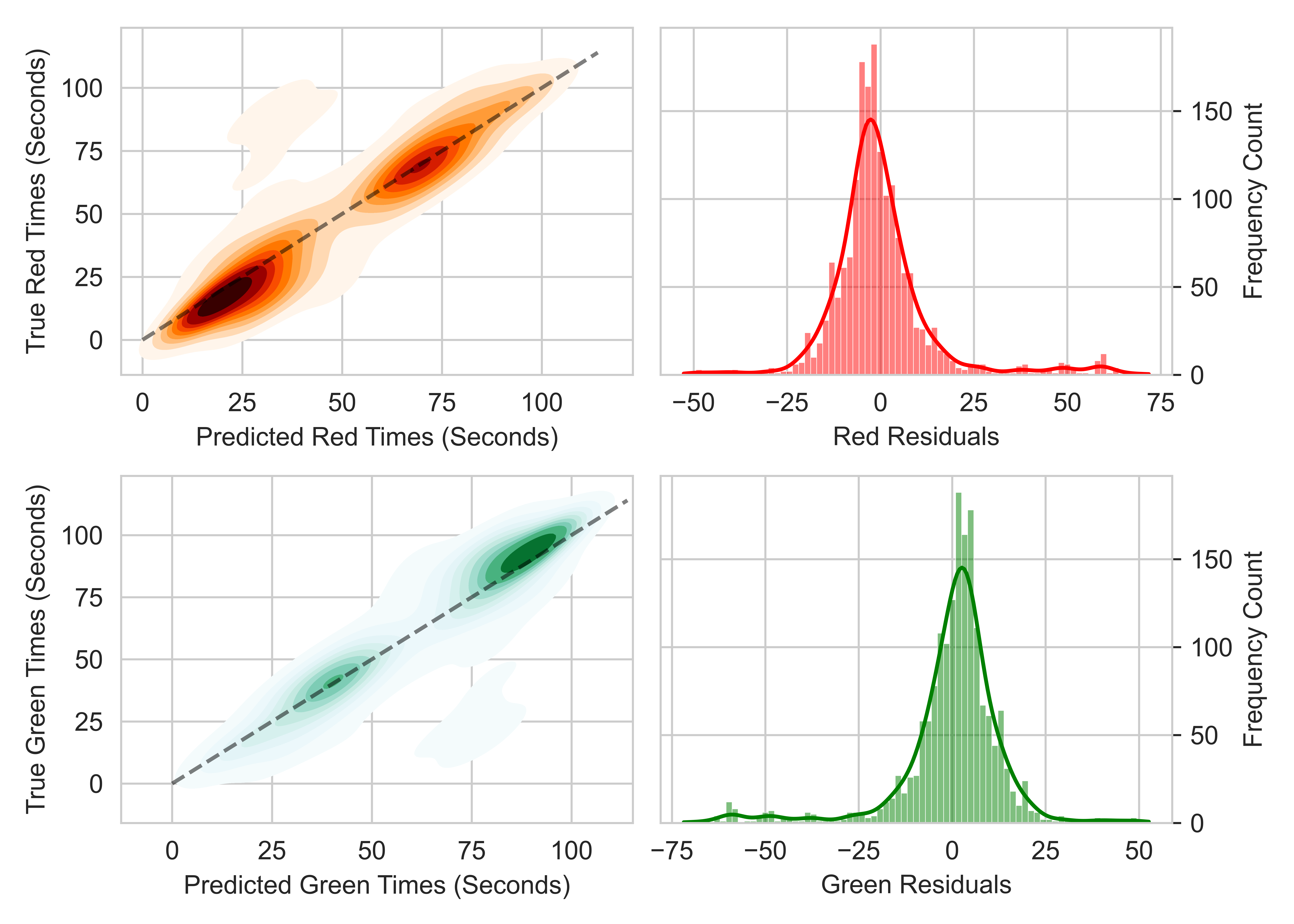}
\caption{Red and Green Times Parity Plots}
\label{fig:red_times}
\end{figure}

\subsection{Red Times and Green Times Results}
After optimizing, the neural network model was able to estimate red time within 7.2 seconds on average with an R-squared score of 0.85. The green times are derived by subtracting the red times from the cycle length for each intersection's phase. We count the yellow time as part of the green time. The accuracy of the green times was very similar to the red times accuracy. In Figure \ref{fig:red_times} we show parity plots for the red and green times on the left side with nearly 2,000 test points. We also include a point density to better understand where the concentration plotted points. Darker shaded areas show the highest concentration of estimates. As you can see, two large concentration of points cluster around the true diagonals. We can also see the spread of error residuals in the histograms on the right side of the Figure \ref{fig:red_times}. This model does moderately well at estimating the red times per phase at each intersection, giving promise to estimating red times with real world probe data. Since this model uses the distribution of stop times over an arbitrary period of time, we do not conduct any penetration rate analysis since we can compensate for a low penetration rate by collecting stop time distributional data for a longer period of time. 

\section{CONCLUSION}

This paper presented an ML approach to estimating  signal timing for pre-timed traffic lights from probe data. We used an XGBoost model to estimate the cycle length and a neural network to estimate the red times per phase. The green times are determined by substracting red times from cycle length durations. Our results demonstrated highly accurate estimations for cycle lengths given only 20 points of data or more and good accurate estimations for red times and green times. 
Our work complements existing literature that has significantly focused on using mathematical methods, such as statistical, probabilistic and optimization based methods, for determining signal timing parameters from vehicle probe data. Future research directions include doing a detailed sensitivity analysis of the performance of the ML-approach given different penetration rates of probe vehicle data, and extending our approach to estimating the signal timing for actuated signals.

\section*{ACKNOWLEDGMENT}

This work was authored by the National Renewable Energy Laboratory, operated by Alliance for Sustainable Energy, LLC, for the U.S. Department of Energy (DOE) under Contract No. DE-AC36-08GO28308. Funding provided by U.S. Department of Energy Vehicle Technology Office. The views expressed in the article do not necessarily represent the views of the DOE or the U.S. Government. The U.S. Government retains and the publisher, by accepting the article for publication, acknowledges that the U.S. Government retains a nonexclusive, paid-up, irrevocable, worldwide license to publish or reproduce the published form of this work, or allow others to do so, for U.S. Government purposes. This work was only made possible through the close cooperation of City of Chattanooga Department of Transportation.

% The authors confirm contribution to the paper as follows: study conception and
% design: Juliette Ugirumurera and Jane Macfarlane; data collection: Joseph Severino and Juliette Ugirumurera; model development: Erik A. Bensen, Joseph Severino, Juliette Ugirumurera,  and Qichao Wang; analysis and
% interpretation of results: Joseph Severino, Qichao Wang, Erik A. Bensen, and Juliette Ugirumurera; draft manuscript
% preparation: Erik A. Bensen, Juliette Ugirumurera, Joseph Severino, and Qichao Wang.

%Bibliography
\bibliographystyle{ieeetr}
\bibliography{bibliography}

\end{document}